\documentclass{article}
\usepackage{authblk}
\usepackage[utf8]{inputenc}
\usepackage{main}
\usepackage{microtype}
\usepackage{subcaption}
\usepackage{graphicx}
\usepackage{times}
\usepackage{latexsym}
\usepackage{amsmath}
\usepackage{float}
\usepackage{footnote}
\usepackage{enumitem}
\usepackage{bm}
\usepackage{arydshln}
\usepackage{booktabs}
\usepackage{array}     
\usepackage{multicol}
\usepackage{multirow}
\usepackage{color}
\usepackage{xcolor}     
\usepackage{colortbl}
\usepackage{bbding}
\usepackage{makecell}
\usepackage{mathtools}
\usepackage{imakeidx}
\usepackage{longtable}
\usepackage{wrapfig}
\usepackage{algorithmic}
\usepackage{rotating}
\makeindex
\usepackage{arydshln}
\usepackage{lipsum}
\usepackage{natbib}
\usepackage[toc]{multitoc}
\usepackage[edges]{forest}
\usepackage[normalem]{ulem}
\definecolor{mydarkblue}{rgb}{0,0.08,0.45}
\usepackage[colorlinks=true,linkcolor=mydarkblue,citecolor=mydarkblue,filecolor=mydarkblue,urlcolor=mydarkblue]{hyperref}
\usepackage{CJKutf8}
\usepackage{awesomebox} 
\usepackage{bbding}
\usepackage[most]{tcolorbox}
\usepackage{booktabs}
\usepackage{geometry}
\definecolor{wkblue}{rgb}{0.2, 0.3, 0.6}
\definecolor{meta-color}{rgb}{0.5, 0.5, 0.5}
\usepackage{amsmath}
\usepackage{enumitem}
\usepackage{lscape} 
\usepackage{booktabs}
\usepackage{algorithm}   
\usepackage{setspace}

\usepackage{algorithmic}
\usepackage{tabularx,booktabs}
\usepackage{makecell}

\usepackage{amssymb}
\usepackage{amsfonts}

\usepackage{multirow}

\usepackage[tikz]{bclogo}
\usepackage[framemethod=tikz]{mdframed}
\definecolor{bgblue}{RGB}{245,243,253}
\definecolor{ttblue}{RGB}{91,194,224}

\usepackage{pgfplots}
\usepackage{pgfplotstable}

\mdfdefinestyle{mystyle}{%
  rightline=true,
  innerleftmargin=10,
  innerrightmargin=10,
  outerlinewidth=3pt,
  topline=false,
  rightline=true,
  bottomline=false,
  skipabove=\topsep,
  skipbelow=\topsep
}

\newtcolorbox{myboxi}[1][]{
  breakable,
  title=#1,
  colback=red!5,
  colbacktitle=red!5,
  coltitle=black,
  fonttitle=\bfseries,
  bottomrule=0pt,
  toprule=0pt,
  leftrule=2pt,
  rightrule=2pt,
  titlerule=0pt,
  arc=0pt,
  outer arc=0pt,
  colframe=red,
}

\newtcolorbox{myboxnote}[1][]{
  breakable,
  title=#1,
  colback=orange!0,
  colbacktitle=orange!0,
  coltitle=black,
  fonttitle=\bfseries,
  bottomrule=0pt,
  toprule=0pt,
  leftrule=2pt,
  rightrule=2pt,
  titlerule=0pt,
  arc=0pt,
  outer arc=0pt,
  colframe=orange,
}

\newtcolorbox{myboxii}[1][]{
  breakable,
  freelance,
  title=#1,
  colback=white,
  colbacktitle=white,
  coltitle=black,
  fonttitle=\bfseries,
  bottomrule=0pt,
  boxrule=0pt,
  colframe=white,
  overlay unbroken and first={
  \draw[red!75!black,line width=3pt]
    ([xshift=5pt]frame.north west) -- 
    (frame.north west) -- 
    (frame.south west);
  \draw[red!75!black,line width=3pt]
    ([xshift=-5pt]frame.north east) -- 
    (frame.north east) -- 
    (frame.south east);
  },
  overlay unbroken app={
  \draw[red!75!black,line width=3pt,line cap=rect]
    (frame.south west) -- 
    ([xshift=5pt]frame.south west);
  \draw[red!75!black,line width=3pt,line cap=rect]
    (frame.south east) -- 
    ([xshift=-5pt]frame.south east);
  },
  overlay middle and last={
  \draw[red!75!black,line width=3pt]
    (frame.north west) -- 
    (frame.south west);
  \draw[red!75!black,line width=3pt]
    (frame.north east) -- 
    (frame.south east);
  },
  overlay last app={
  \draw[red!75!black,line width=3pt,line cap=rect]
    (frame.south west) --
    ([xshift=5pt]frame.south west);
  \draw[red!75!black,line width=3pt,line cap=rect]
    (frame.south east) --
    ([xshift=-5pt]frame.south east);
  },
}

\usepackage{fancyhdr} 
\usepackage{blindtext} 
\usepackage{makecell}

\DeclareCaptionFont{black}{\color{black}}

\definecolor{myblue}{rgb}{0.9, 0.1, 0.94}
\definecolor{mygreen}{rgb}{0.64, 0.56, 0.88}
\definecolor{myyellow}{rgb}{0.68, 0.6, 0.1}
\definecolor{fancygreen}{rgb}{0.33, 0.68, 0.20}
\definecolor{salmon}{rgb}{0.94, 0.52, 0.49}
\definecolor{tablegreen}{rgb}{0.82, 0.94, 0.75}
\definecolor{tableblue}{rgb}{0.81, 0.90, 0.94}
\definecolor{tablered}{rgb}{0.97, 0.85, 0.85}
\definecolor{tableorange}{rgb}{0.96, 0.85, 0.81}

\newenvironment{itemize*}%
 {\leftmargini=10pt\begin{itemize}%
  \setlength{\itemsep}{0pt}%
  \setlength{\parskip}{0pt}%
  }%
 {\end{itemize}}
\newenvironment{enumerate*}%
 {\begin{enumerate}%
  \setlength{\itemsep}{0pt}%
  \setlength{\parskip}{0pt}}%
 {\end{enumerate}}

\usepackage{xcolor}
\usepackage{listings}  

\newcommand\JSONnumbervaluestyle{\color{blue}}
\newcommand\JSONstringvaluestyle{\color{red}}

\newif\ifcolonfoundonthisline

\makeatletter

\lstdefinestyle{json}
{
  showstringspaces    = false,
  keywords            = {false,true},
  alsoletter          = 0123456789.,
  morestring          = [s]{"}{"},
  stringstyle         = \ifcolonfoundonthisline\JSONstringvaluestyle\fi,
  MoreSelectCharTable =%
    \lst@DefSaveDef{`:}\colon@json{\processColon@json},
  basicstyle          = \ttfamily,
  keywordstyle        = \ttfamily\bfseries,
}

\newcommand\processColon@json{%
  \colon@json%
  \ifnum\lst@mode=\lst@Pmode%
    \global\colonfoundonthislinetrue%
  \fi
}

\lst@AddToHook{Output}{%
  \ifcolonfoundonthisline%
    \ifnum\lst@mode=\lst@Pmode%
      \def\lst@thestyle{\JSONnumbervaluestyle}%
    \fi
  \fi
  \lsthk@DetectKeywords%
}

\lst@AddToHook{EOL}%
  {\global\colonfoundonthislinefalse}

\makeatother

\usepackage{etoolbox}
\usepackage{natbib}
\usepackage{url}
\newcounter{bibcount}
\makeatletter
\patchcmd{\@lbibitem}{\item[}{\item[\hfil\stepcounter{bibcount}{[\thebibcount]}}{}{}
\renewcommand\NAT@bibsetup%
  [1]{\setlength{\leftmargin}{\bibhang}\setlength{\itemindent}{-\parindent}%
      \setlength{\itemsep}{\bibsep}\setlength{\parsep}{\z@}}
\makeatother

\definecolor{mybrown}{RGB}{128,64,0}

\definecolor{titlecolor}{HTML}{4c9cff}

\begin{document}


\title{ResearcherBench: Evaluating Deep AI Research Systems \\ on the Frontiers of Scientific Inquiry}

\author[1,3]{Tianze Xu\textsuperscript{*}}
\author[1,3]{Pengrui Lu\textsuperscript{*}}
\author[1,3]{Lyumanshan Ye}
\author[2,3]{Xiangkun Hu}
\author[1,2,3]{Pengfei Liu\textsuperscript{†}}
\affil{Shanghai Jiao Tong University \quad \textsuperscript{2}SII \quad \textsuperscript{3}GAIR}
  
\maketitle
\pagestyle{fancy}
\thispagestyle{fancy}
\fancyhead{}
\lhead{
  \raisebox{-0.3cm}{\includegraphics[height=0.95cm]{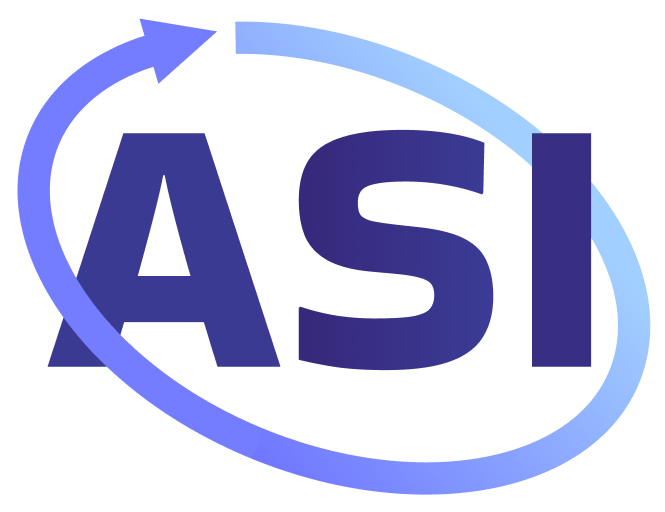}}
}
\rhead{%
  \raisebox{-0.2cm}{\includegraphics[height=0.7cm]{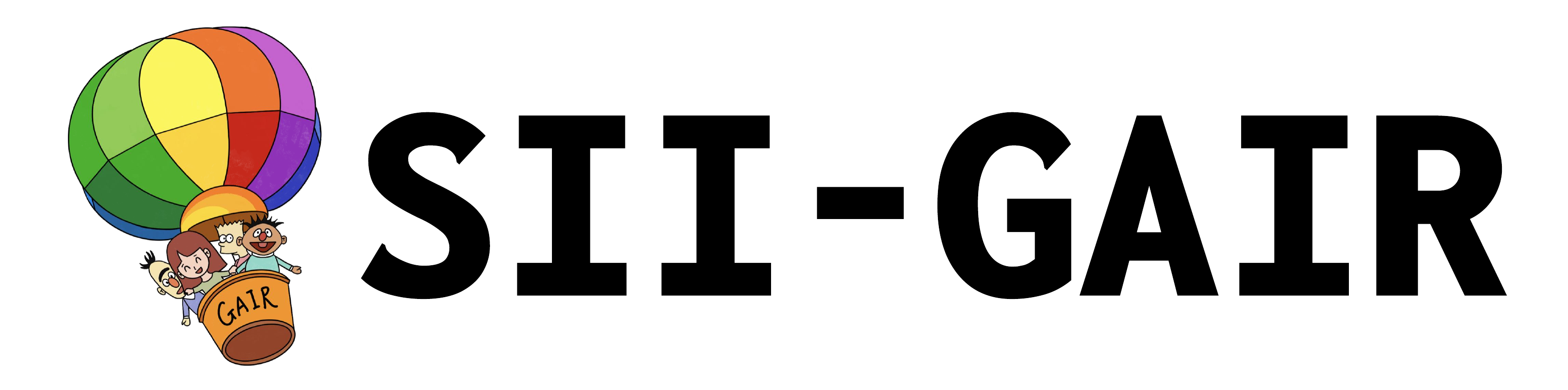}}%
}
\renewcommand{\headrulewidth}{0pt}
\setlength{\headsep}{2mm} 


\footnotetext[1]{* Equal contribution.}
\footnotetext[2]{† Corresponding author.}

\begin{abstract}

The emergence of deep research systems presents significant capabilities in problem-solving, extending from basic queries to sophisticated research tasks. However, existing benchmarks primarily evaluate these systems as agents for web retrieval and report generation, overlooking their potential to discover novel insights on the frontiers of scientific research. To address this gap, we introduce ResearcherBench, the first benchmark focused on evaluating the capabilities of these advanced, agentic systems — which we refer to as Deep AI Research Systems (DARS) — on frontier AI scientific questions. We compiled a dataset of 65 research questions expertly selected from real-world scientific scenarios such as laboratory discussions and interviews, spanning 35 different AI subjects and categorized into three types: technical details, literature review, and open consulting. Our dual evaluation framework combines rubric assessment, which uses expert-designed criteria to evaluate insight quality, with factual assessment, which measures citation accuracy (faithfulness) and coverage (groundedness). We evaluated several leading commercial DARS and baseline systems. Results show that OpenAI Deep Research and Gemini Deep Research significantly outperform other systems, with particular strength in open-ended consulting questions. Such capabilities represent a meaningful step toward AI self-improvement, aligning with the vision of ASI for AI. We open-source ResearcherBench to provide a standardized platform for promoting the development of next-generation AI research assistants, hoping to foster a new perspective in AI research evaluation for a novel pattern of scientific collaboration: \url{https://github.com/GAIR-NLP/ResearcherBench}.
\end{abstract}

\vspace{20px}

\begin{figure}[htp]
\centering
\includegraphics[width=\textwidth]{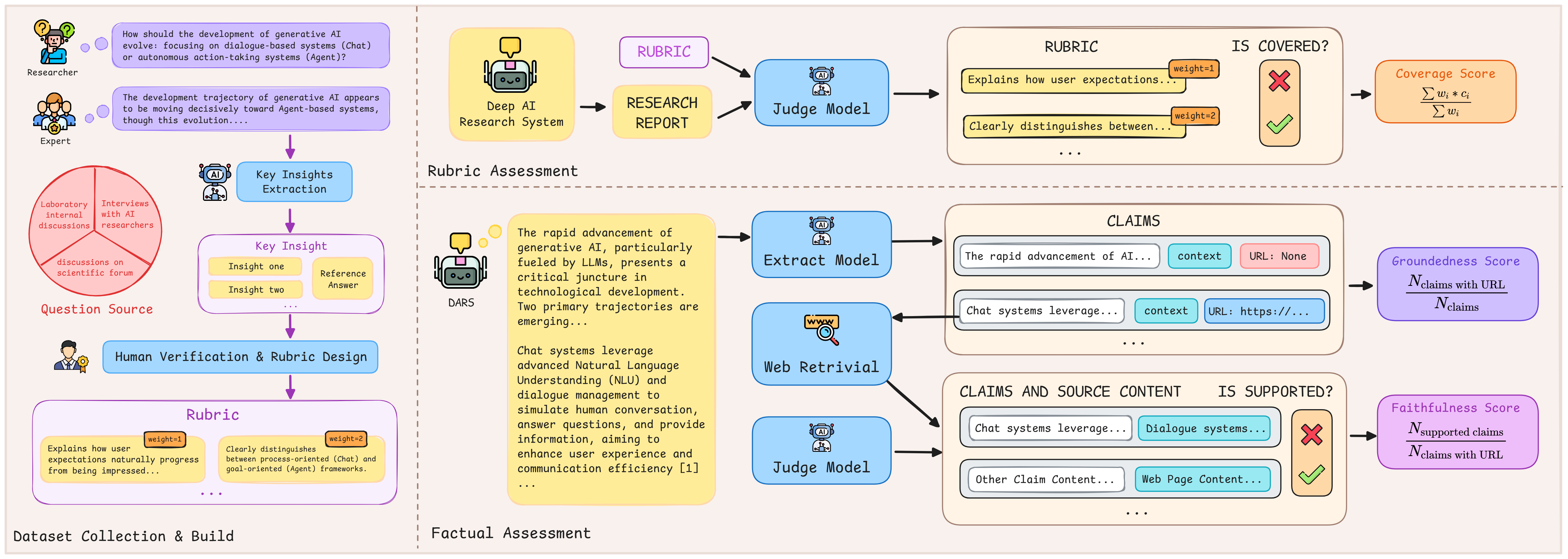}
\caption{\textbf{ResearcherBench Framework Overview.} The framework consists of three main components from top to bottom: (1) Dataset collection from authentic research scenarios leading to expert-generated rubrics, (2) Rubric assessment to evaluate coverage against rubrics, and (3) Factual assessment to measure faithfulness and groundedness scores.}
\label{fig:overview}
\end{figure}

\newpage

\pagestyle{fancy}
\lhead{\rightmark}
\renewcommand{\headrulewidth}{0.7pt}
\setlength{\headsep}{5mm}

\clearpage

\newpage

\section{Introduction}

The advent of artificial intelligence has fundamentally transformed how we approach complex problem-solving tasks \cite{gridach2025agenticaiscientificdiscovery}, with deep research systems emerging as sophisticated AI agents capable of autonomously conducting intricate research workflows \cite{zheng2025deepresearcher, huang2025deepresearchagentssystematic}. These systems integrate advanced information retrieval, analysis, and synthesis capabilities, demonstrating remarkable proficiency in processing vast amounts of information and generating comprehensive reports across diverse domains \cite{xu2025comprehensivesurveydeepresearch}.

Deep research systems have also been increasingly deployed by researchers to assist with various aspects of scientific inquiry, such as investigating technical implementation details, conducting comprehensive literature reviews, and synthesizing existing knowledge \cite{hai2025aiindex}. These applications have demonstrated clear value in streamlining traditional research workflows and enhancing productivity in well-established research domains. However, the scientific community might overlook a potentially transformative capability of these systems: their potential to assist researchers in exploring genuinely open-ended, frontier questions that exist at the cutting edge of scientific knowledge \cite{lu2024aiscientistfullyautomated,google_ai_coscientist_2025}.

This transition from being systems that merely retrieve and summarize information to becoming genuine ``research partners'' capable of valuable collaboration on unexplored scientific territories represents one of the most significant challenges facing the development of AI research assistants today \cite{wang2024humanaimutuallearningnew}. We define this emerging category as \textbf{Deep AI Research Systems (DARS)}, the sophisticated agentic systems capable of dynamic reasoning and autonomously conducting intricate research workflows with multi-iteration web retrieval and tool uses~\cite{huang2025deepresearchagentssystematic}. The capabilities of such systems points toward a profound objective: by systematically involving DARS in challenging frontier AI research, we can create a powerful feedback loop for recursive self-improvement, where AI is used to accelerate its own development, aligning with the broader vision of Artificial Superintelligence (ASI) for AI.

However, this raises a critical question: \textbf{can current DARS truly assist human researchers in tackling the most challenging, high-valued, and open-ended questions at the frontiers of science}, where definitive answers do not yet exist and novel insights must be synthesized from fragmentary and cross-domain information? 

Existing benchmarks for evaluating deep research capabilities predominantly focus on assessing systems' abilities to retrieve and synthesize established knowledge rather than their capacity to engage with genuinely novel, frontier research questions. Current evaluation frameworks typically emphasize comprehensive report generation \cite{du2025deepresearch} or agents' web interaction capabilities \cite{gou2025mind2web, openai_browsecomp}, focusing on breadth of information retrieval rather than conceptual understanding and insight generation. These frameworks fail to capture a crucial dimension of research assistance: the ability to understand, analyze, and provide meaningful insights on highly specialized, cutting-edge scientific problems. \cite{zheng2025deepresearcher, starace2025paperbench} Such problems are characterized by inherent ambiguity, absent definitive answers, and the need for creative synthesis of disparate ideas \cite{alzubi2025opendeepresearch}.

To address this critical gap in evaluation methodology, we introduce \textbf{ResearcherBench}, the first benchmark specifically designed to evaluate DARS capabilities on frontier scientific questions, as shown in Figure \ref{fig:overview}. Unlike existing benchmarks that primarily assess information retrieval and general synthesis abilities, ResearcherBench focuses on evaluating whether AI systems can provide meaningful assistance to human researchers working on genuinely unsolved, cutting-edge problems in the field of artificial intelligence.

Our benchmark represents a paradigm shift in evaluation philosophy—moving from assessing ``whether deep research systems can retrieve and summarize information'' to evaluating ``whether DARS can understand complex problems and provide meaningful insights as genuine research partners.'' This approach recognizes that true research assistance requires not just information gathering, but deep comprehension of nuanced concepts, the ability to deeply explore connections between different perspectives, and the capacity to generate novel insights that advance scientific understanding.

ResearcherBench makes several significant contributions to the field of AI research evaluation:

\begin{itemize}
    \item \textbf{A Novel Task Collection:} We present a carefully curated dataset of 65 high-quality research questions sourced from authentic frontier scenarios, including laboratory discussions, interviews with leading AI researchers, and active scientific forums. These questions span 35 distinct AI research subjects and are categorized into three distinct types. This categorization enables nuanced evaluation of DARS capabilities across different types of research assistance scenarios.
    \item \textbf{A Unique Dual Evaluation Framework:} Our assessment methodology combines rubric assessment with factual assessment. The rubric assessment employs domain expert-crafted evaluation criteria tailored to each specific question, ensuring alignment with human-anchored high-value insights. The factual assessment framework introduces two complementary metrics: faithfulness score and and groundedness score, to evaluate the overall factual accuracy and coverage of generated content.
    \item \textbf{Comprehensive Empirical Analysis:} Our extensive evaluation of leading commercial DARS platforms provides a holistic, multi-faceted benchmark of their capabilities and fundamental limitations. The analysis reveals their primary strength in exploratory reasoning for open-ended tasks, rather than precise technical or literature synthesis. The evaluation also uncovers a paradoxical ``high faithfulness, low groundedness'' pattern, and shows that high citation coverage does not necessarily equate to superior insight quality. These findings from analysis validate DARS' potential as innovative research partners.
    \item \textbf{Open-Source Contribution:} We are releasing ResearcherBench as a comprehensive evaluation platform, encompassing our curated dataset of frontier questions and the dual evaluation framework. This initiative provides the community with a standardized infrastructure to benchmark DARS capabilities of AI researching, to collaboratively advance the development of AI systems capable of valuable scientific research assistance.
\end{itemize}

\section{Related Work}

\subsection{Deep AI Research Systems}

Deep AI Research Systems (DARS) represent a significant evolution in AI research capabilities, designed to conduct comprehensive research by autonomously retrieving, analyzing, and synthesizing information from diverse sources. Unlike traditional retrieval-augmented generation (RAG)~\cite{gao2024retrievalaugmentedgenerationlargelanguage} systems that operate within constrained local databases, DARS are sophisticated agentic systems capable of dynamic reasoning, adaptive planning, multi-iteration external data retrieval on the open web for complex tasks~\cite{huang2025deepresearchagentssystematic}.

The evolution of AI research systems has progressed from basic prompt-based approaches \cite{zheng2024openresearcher, alzubi2025opendeepresearch} to more sophisticated frameworks incorporating supervised fine-tuning (SFT) \cite{asai2024openscholar} and reinforcement learning methods \cite{jin2025searchr1, chen2025ReSearch, song2025r1searcher}. However, these earlier systems faced fundamental limitations due to their reliance on static knowledge repositories and lack of iterative reasoning capabilities \cite{zheng2025deepresearcher}.

Modern DARS address these limitations through autonomous query formulation, iterative content reflection, and dynamic search refinement processes. Recent work \cite{zheng2025deepresearcher, jin2025searchr1trainingllmsreason} pioneered comprehensive end-to-end training frameworks for LLM-based deep research agents, implementing specialized multi-agent architectures for authentic web search interactions. Leading commercial implementations \cite{openai_deep_research, gemini_deep_research, xai_grok_3, perplexity_deep_research} demonstrate advanced capabilities including multi-step web browsing and synthesis, iterative multi-point research planning, truth-seeking across vast knowledge corpora, and comprehensive report generation with iterative research processes.

\subsection{Research-related Evaluation Benchmarks}

The evaluation of AI research capabilities has evolved through distinct methodological frameworks, each targeting different aspects of research proficiency. Traditional RAG-based benchmarks \cite{joshi2017triviaqa, yang2018hotpotqa} established foundations for multi-hop reasoning evaluation, while more recent work \cite{asai2024openscholar, woodrow2025algorithms} advanced literature synthesis assessment with expert-written responses across scientific disciplines, emphasizing citation accuracy and factual correctness.

Recent developments have introduced benchmarks specifically designed for evaluating deep research capabilities. DeepResearch Bench \cite{du2025deepresearch} presents 100 PhD-level research tasks crafted by domain experts across 22 distinct fields, introducing RACE (Reference-based Adaptive Criteria-driven Evaluation) and FACT (Framework for Factual Abundance and Citation Trustworthiness) methodologies to assess report generation quality and information retrieval capabilities. Mind2Web 2 \cite{gou2025mind2web} focuses on agentic search evaluation with 130 realistic, long-horizon tasks requiring real-time web browsing and information synthesis, introducing an Agent-as-a-Judge framework for automated assessment of answer correctness and source attribution. 

Additionally, OpenAI has also proposed related benchmarks for deep research, with evaluation frameworks like Browsecomp~\cite{openai_browsecomp} that measure persistent web browsing abilities through questions requiring navigation to locate complex, entangled information, and specialized benchmarks like PaperBench~\cite{starace2025paperbench} evaluate research replication capabilities across academic papers with individually gradable tasks.

\vspace{8pt}

Despite these advances, existing benchmarks focus on assessing established technical competencies and knowledge synthesis, failing to evaluate the nuanced capabilities required for genuinely frontier research questions. For example, while adaptive criteria generated by large language models~\cite{du2025deepresearch} may be suitable for general research reports, the evaluation rubric for frontier research problems cannot be reliably generated directly by them, as this task requires extensive domain expertise and a nuanced understanding of what constitutes valuable insights in cutting-edge research. This gap motivates the need for evaluation frameworks that can assess AI systems' potential as genuine research partners in exploring uncharted scientific territories, where novel insight generation and deep conceptual understanding are paramount.

\section{ResearcherBench}

\begin{figure}[htbp]
\centering
\includegraphics[width=\textwidth]{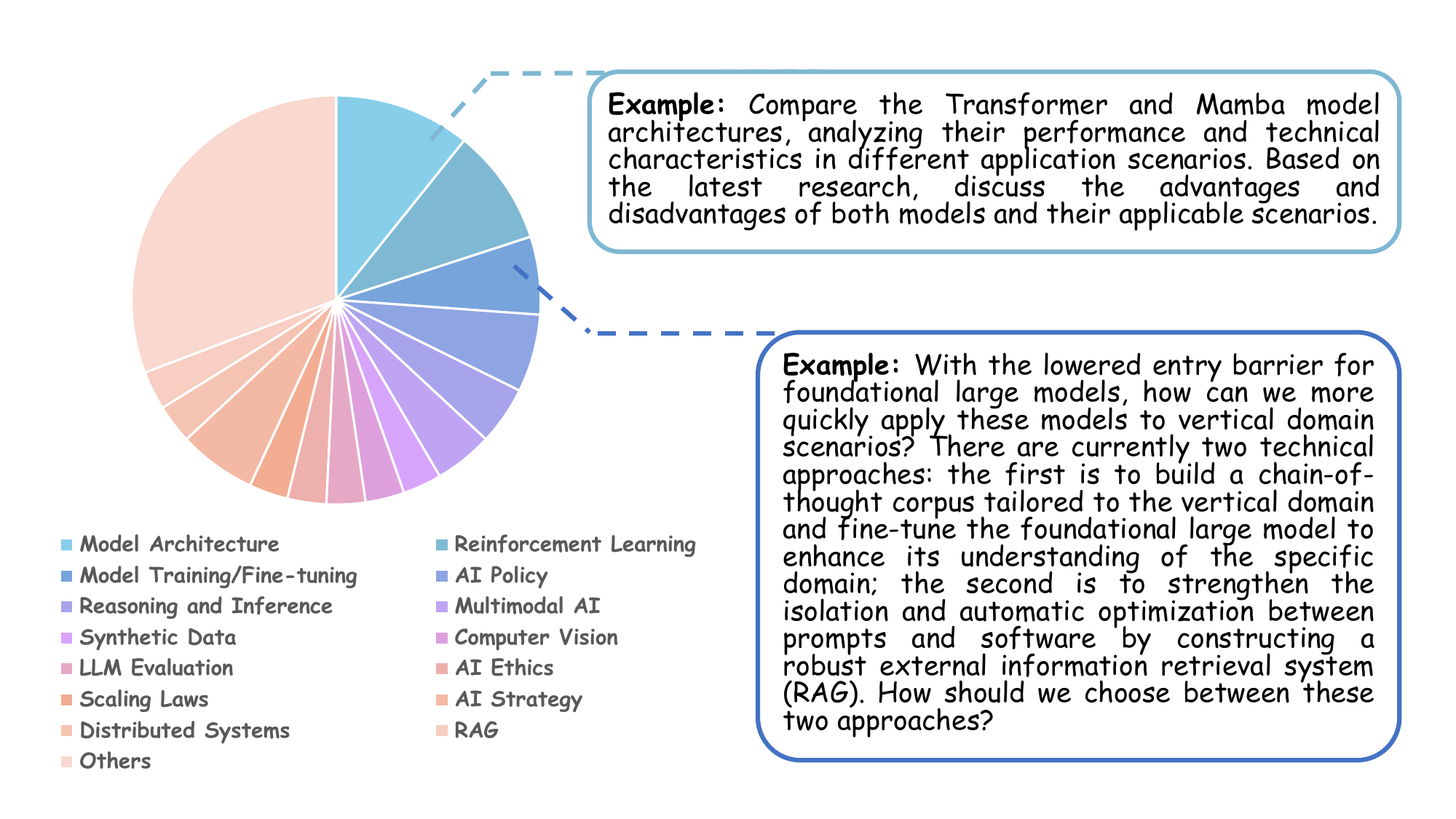}
\caption{\textbf{AI Benchmark Topic Distribution with Representative Examples.} Left Side: Pie chart showing the distribution of AI subjects in the benchmark. Right Side: Concrete question examples from major subjects.}
\label{fig:distribute}
\end{figure}

ResearcherBench presents a systematic approach to constructing a comprehensive benchmark for evaluating DARS capabilities on frontier research questions. We employed rigorous data collection and filtering methodologies to curate authentic research questions from real-world scientific scenarios, resulting in a high-quality dataset of 65 questions across 35 AI subjects, as shown in Figure \ref{fig:distribute}.

\subsection{Data Collection Strategy}

Our benchmark construction follows a systematic approach designed to capture authentic frontier research questions from real-world scientific scenarios. We identified three primary contexts that consistently and naturally generate high-quality frontier research questions: (1) \textbf{Laboratory internal research discussions}, where researchers actively grapple with unsolved technical challenges; (2) \textbf{Interviews with leading AI researchers}, which often reveal emerging research directions and open problems; and (3) \textbf{Scientific forum discussions}, where practitioners discuss implementation challenges and theoretical gaps.

We systematically collected research questions from these authentic scenarios, resulting in an initial corpus of several hundred candidate questions. Our collection methodology prioritized questions that emerged organically from research practice rather than artificially constructed queries, ensuring data validity of our benchmark.

\subsection{Dataset Composition and Categorization}

We utilized \texttt{Claude-3.7-Sonnet} to systematically classify our questions across research domains and question types. The questions were categorized into three distinct types: technical details, literature review, and open consulting, as shown in Table \ref{tab:q_select}. This categorization enables us to systematically evaluate DARS capabilities across different research assistance scenarios and better highlight how these systems perform when facing diverse cognitive demands.

Our classification also resulted in coverage of 35 distinct AI research subjects including model architecture, multimodal fusion, AI ethics, and emerging paradigms in machine learning. This comprehensive topical distribution ensures that our benchmark captures the majority of scenarios in frontier AI research, providing a representative evaluation framework for assessing DARS capabilities across the full spectrum of contemporary AI research challenges. Details of subject distribution visualization and examples are shown in Figure \ref{fig:distribute}.

\subsection{Question Selection and Filtering}

To filter high-quality research questions from the initial collection of several hundred research problems for constructing our benchmark, we designed a detailed question selection framework across several dimensions including quality, clarity, and verifiability, etc. based on different question types. Detailed criteria and specific rubrics for each question type are provided in Appendix \ref{app:selection_criteria}.

We engaged experienced AI domain researchers and practitioners as annotators to evaluate candidate questions against these criteria. Through this systematic review process, we refined the initial corpus to a final set of 65 research questions that met our stringent selection standards, resulting in ResearcherBench's final dataset distributed across the three question types and 35 specialized AI research subjects.

\begin{table}[h]
    \centering
    \renewcommand{\arraystretch}{1.3} 
    \begin{tabularx}{\textwidth}{>{\centering\arraybackslash}l>{\raggedright\arraybackslash}p{4.5cm}>{\raggedright\arraybackslash}X}
        \toprule
        \textbf{Question Type} & \textbf{Definition} & \textbf{Example}  \\
        \midrule
        \textbf{Technical Details} & Questions requiring explanations of methodologies, implementations, or theoretical concepts with a strong emphasis on accuracy and verification. & \texttt{How can we improve large language models' effectiveness on long text reasoning tasks (such as fact extraction and summarization) and avoid the phenomenon where key information is easily overlooked in long contexts? Answer from the perspectives of model architecture, training methods, inference strategies, and model evaluation.}  \\
        \addlinespace
        \textbf{Literature Review} & Questions that involve synthesizing findings from multiple research papers, comparing methodologies, and identifying trends or gaps in existing literature. & \texttt{For complex reasoning tasks (e.g., tasks involving multiple citations or extended reasoning chains), what are the strengths of current agent technologies, and what are their limitations? Please analyze this in the context of research since June 2024.} \\
        \addlinespace
        \textbf{Open Consulting} & Questions that explore emerging trends, strategic insights, detailed solutions or broader implications, often requiring subjective interpretation and expert judgment. & \texttt{Could transformer architectures be fundamentally reimagined to process multimodal inputs (e.g. video, audio, or text) with the same efficiency they process text?}  \\
        \addlinespace
        \bottomrule
    \end{tabularx}
    \caption{\textbf{Question Types.} Each question type follows a detailed definition and an example. Technical detail questions emphasize precise verification, literature review questions focus on comprehensive synthesis, and open consulting questions prioritize broader insights.}
    \label{tab:q_select}
\end{table}

\section{Dual Evaluation Framework}
\label{sec:Evaluation_Methodology}
ResearcherBench introduces a dual evaluation framework that comprehensively assesses DARS performance through both rubric assessment and factual assessment. We combine expert-designed criteria evaluation with automated factual verification to capture both depth of understanding and reliability of generated research reports.

\subsection{Rubric Assessment: Expert-designed Criteria Evaluation}


Evaluating responses to frontier research questions poses unique challenges that cannot be adequately addressed through simple LLM-as-a-Judge approaches~\cite{gu2025surveyllmasajudge,li2025generationjudgmentopportunitieschallenges}. The nuanced nature of cutting-edge research requires assessment of conceptual depth, theoretical understanding, and insight quality—dimensions that are difficult to capture through binary or holistic scoring methods. 

Our fine-grained rubric assessment framework addresses these limitations by decomposing complex research questions into multiple specific, evaluable components that collectively capture the essential dimensions of high-quality scholarly responses. Each rubric consists of expert-designed evaluation criteria with assigned importance weights, focusing on conceptual understanding, methodological rigor, and analytical depth rather than mere factual recall, thereby enabling comprehensive assessment of research insight quality.

\subsubsection{Criteria Construction}

The evaluation criteria for frontier research problems cannot be reliably generated directly through large language models, as they require extensive domain expertise and nuanced understanding of what constitutes valuable insights in cutting-edge research \cite{dorner2024limits}. Therefore, human experts must participate in designing these criteria to ensure they accurately reflect the standards and expectations of frontier scientific inquiry. We design a three-step method to construct these rubrics:

\paragraph{Insight Extraction.}

We firstly use \texttt{Claude-3.7-Sonnet} to analyze and extract key insights from multiple diverse contextual sources for each question, including original discussion records, relevant academic literature, expert opinions and insights, technical background materials, cross-disciplinary reference materials, and industry practice cases. This integration process generates comprehensive reference materials that contain these high-value insights, which then serve as auxiliary materials for human expert annotation. Domain experts subsequently review and validate these integrated references to ensure accuracy and completeness, while supplementing any professional insights and judgments that the system might have missed from the multi-source synthesis, establishing a solid foundation for subsequent rubric development.


\paragraph{Rubric Item Design.}
Following established guidelines and templates, we invited human annotators (experienced masters, Ph.D.s or professionals majored in AI) to transform these extracted key insights into operational evaluation rubric. Each rubric item represents a specific aspect of reasoning or insight that should be addressed in a comprehensive response. Annotators also assigned weights to every rubric item based on their relative importance and impact on overall answer quality. Detailed guidelines for rubric construction are provided in Appendix~\ref{app:rubric_guidelines}.

\paragraph{Quality Control.} To ensure rubric reliability and validity, we implemented a multi-stage quality control process. Each rubric was collaboratively developed by two experienced AI researchers, with one researcher responsible for initial drafting and weight assignment, and the other conducting comprehensive review and collaborative refinement. All rubrics underwent expert review to assess completeness, clarity, independence, and discriminative power. We conducted pilot testing with sample DARS responses to identify and refine problematic rubric items through iterative revision. Finally, we validated rubric effectiveness through meta-evaluation comparing human expert judgments with automated assessment, as detailed in Section \ref{sec: judge_model_sel}.

\subsubsection{Evaluation Metric}

For each question, we evaluate whether DARS responses cover the key insights specified in the expert-designed rubric. Let $Q_k$ denote the $k$-th question in our benchmark, and $\text{Res} = \text{DARS}(Q_k)$ represent the final response generated by the DARS system. The binary indicator $c_i$ for each rubric item is computed as:

$$
c_i = \text{Judge}(Q_k, \text{Res}, r_i)
$$

where $r_i$ represents the content of the $i$-th rubric item, and the judge model returns 1 if the response meets the criteria of this rubric item, and 0 otherwise. The result of such rubric-based assessment, which we term as \emph{coverage score}, is calculated as:

\begin{equation}
\text{Coverage Score} = \frac{\sum_{i=1}^{n} w_i \cdot c_i}{\sum_{i=1}^{n} w_i}
\end{equation}

where $w_i$ represents the weight of the $i$-th rubric item, and $n$ is the total number of rubric items. This formulation considers both the coverage of the expert-aligned criteria and the importance of each rubric item.

\subsection{Factual Assessment: Faithfulness and Groundedness Evaluation}


While rubric assessment evaluates insight quality and conceptual depth, the factuality of DARS-generated research reports remains a fundamental requirement for reliability~\cite{zhang2023siren}. Our factual assessment framework addresses the specific challenges of evaluating citation accuracy and groundedness~\cite{ragas2024} in DARS-generated research reports. 

\subsubsection{Assessment Methodology}

The factual assessment framework is a three-step process that automatically evaluates the citation accuracy and content groundedness of DARS-generated research reports:

\paragraph{Claim Extraction.}
We employ an extract model to extract all factual claims within DARS-generated reports along with their corresponding context passages. The extract model also examines whether each claim can be retrieved with a corresponding citation URL from the report. If a claim can be linked with a citation URL, we save it as a URL-claim-context triplet for subsequent verification. Otherwise, this claim is considered ungrounded, and its URL is marked as empty, indicating the absence of explicit source attribution for this factual assertion.

Let $Q_k$ denote the $k$-th question in our benchmark, and $\text{Res}, \text{Ref} = \text{DARS}(Q_k)$ represent the main content and reference section of the research report generated by the DARS system. For each $Q_k \in Q$, we denote $C_k$ is the set of all claims for question $Q_k$, which can be represented as: 

$$C_k = \{c_i = (\text{text}_i, \text{context}_i, \text{url}_i) \mid c_i = \text{Extract}(Q_k, \text{Res}, \text{Ref}), i = 1, 2, ..., N_k\}$$

where $\text{text}_i$ is the textual content of the claim extracted from responses, $\text{url}_i$ is the corresponding URL extracted from references if exists ($url_i \in \{\text{URL} \cup \{\emptyset\}\}$), $\text{context}_i$ is the context surrounding the claim used as supplementary information for verification, and $N_k = |C_k|$ is the total number of claims for $Q_k$.

\paragraph{Citation Support Verification.}
For each URL-claim-context triplet, we extract textual content from the URL sources using the Jina Reader API.\footnote{\url{https://jina.ai/reader}} Then we use a judge model to perform binary evaluation of whether the extracted content supports the corresponding claim. When the extracted claim is semantically incomplete or ambiguous to judge, the context passage can serve as supplementary information to assist the model's judgment. It finally outputs a binary result (`yes' or `no') for each triplet.

We denote that $C^{cited}_k \subseteq C_k$ is the subset of cited claims with non-empty URLs, and $C^{supp}_k \subseteq C^{cited}_k$ is the subset of claims that are both cited and supported by their URL sources. They can be represented as:

\begin{align*}
    C^{cited}_k &= \{c_i = (\text{text}_i, \text{context}_i, \text{url}_i) \mid c_i \in C_k \text{ and } url_i \neq \emptyset\} \\
C^{supp}_k &= \{c_i \in C^{cited}_k \mid \text{Judge}(\text{text}_i, \text{context}_i, \text{SourceText}(\text{url}_i)) = 1\}
\end{align*}

where $\text{SourceText}(\text{URL})$ means the textual content extracted from the URL, and $\text{Judge}(\text{text}, \text{context}, \text{SourceText}(\text{URL}))$ returns 1 if the claim surrounded in the context is supported by URL, and 0 otherwise.

\paragraph{Score Computation.}
Based on verification results, we calculate two complementary metrics to assess the overall factual reliability of DARS-generated report: 
\begin{itemize}
    \item \textbf{Faithfulness score}, which measures the accuracy of citations in supporting their corresponding claims. This metric evaluates the proportion of cited claims that are actually supported by their referenced sources, indicating the reliability of the citation-claim relationships when citations are provided.
    \item \textbf{Groundedness score}, which evaluates the overall citation coverage of response content. This metric measures the proportion of all factual claims that have explicit citation support, reflecting how comprehensively the research report grounds its assertions in verifiable sources rather than relying on unsupported statements.
\end{itemize}
 
 Suppose that $N_{c,k} = |C^{cited}_k|$ is the total number of cited claims, and $N_{s,k} = |C^{supp}_k|$ is the total number of supported claims, these two metric are calculated as:

\begin{equation}
\text{Faithfulness Score} = \frac{N_{s,k}}{N_{c,k}}
\end{equation}

\begin{equation}
\text{Groundedness Score} = \frac{N_{c,k}}{N_{k}}
\end{equation}

\vspace{10px}

ResearcherBench's dual evaluation framework that assesses DARS capabilities across two distinct dimensions: rubric-based evaluation for insight quality and conceptual depth, and factual assessment for citation accuracy and content groundedness. This framework allows for systematic evaluation of both the intellectual contribution and empirical reliability of DARS performance on frontier research tasks, providing a holistic view of system performance on frontier research tasks.

\section{Experiments and Results}
\subsection{Experimental Setup}
\paragraph{Evaluated Systems.} 
We evaluated several leading commercial deep research systems to assess their performance on frontier AI research questions: OpenAI Deep Research \cite{openai_deep_research}, Gemini Deep Research powered by Gemini-2.5-Pro~\cite{gemini_deep_research}, Grok DeepSearch \& DeeperSearch~\cite{xai_grok_3} and Perplexity Deep Research \cite{perplexity_deep_research}. To provide comprehensive comparison, we also evaluated LLM systems with web search capabilities: GPT-4o Search Preview, Perplexity: Sonar Reasoning Pro. Details on our specific interaction procedures with each system can be found in Appendix \ref{app:experiment_setup}.

\paragraph{Evaluation Configuration.} 

After comprehensive evaluation of multiple candidate Judge LLMs (detailed in Section \ref{sec: judge_model_sel}), we selected \texttt{o3-mini} as the judge model for rubric assessment to evaluate rubric item coverage. For factual assessment, we chose \texttt{GPT-4.1} as both the extraction model and judge model for its superior context length and factual verification accuracy. All evaluations were conducted between March and April in 2025 to ensure temporal consistency and fair comparison across systems, with detailed data collection procedures provided in Appendix~\ref{app:data_coll}. 

\paragraph{Implementation Details.}

Our experimental evaluation pipeline was constructed based on the framework outlined in Section \ref{sec:Evaluation_Methodology}. However, in practice, we made several minor adjustments considering exception handling and robustness requirements. These implementation-specific details including prompts used in the evaluation framework can be found in Appendix~\ref{app:implement_details}.

\subsection{Main Results and Findings}

\begin{table}[htbp]
\centering

\begin{tabular}{lccc}
\toprule
\multirow{2}{*}{\textbf{Model}} & \textbf{Rubric Assessment} & \multicolumn{2}{c}{\textbf{Factual Assessment}} \\
\cmidrule(lr){2-2} \cmidrule(lr){3-4}
 & \textbf{Coverage} & \textbf{Faithfulness} & \textbf{Groundedness} \\
\midrule
\multicolumn{4}{c}{\textit{Deep Research System}} \\
\midrule
OpenAI Deep Research & \textbf{0.7032} & 0.84 & 0.34 \\
Gemini Deep Research & \underline{0.6929} & \textbf{0.86} & \underline{0.59} \\
Grok3 DeepSearch & 0.4414 & 0.69 & 0.32 \\
Grok3 DeeperSearch & 0.4398 & 0.80 & 0.31 \\
Perplexity Deep Research & 0.4800 & \underline{0.85} & 0.56 \\
\midrule
\multicolumn{4}{c}{\textit{LLM with Search Tools}} \\
\midrule
GPT-4o Search Preview & 0.3576 & \textbf{0.86} & 0.39 \\
Perplexity: Sonar Reasoning Pro & 0.4663 & 0.62 & \textbf{0.68} \\
\bottomrule
\end{tabular}
\caption{Comprehensive Evaluation Results across Different Assessment Metrics. \textbf{Bold values} indicate the best performance for each metric, while \underline{underlined values} represent the second-best performance.}
\label{tab:results}
\end{table}

\subsubsection{Overview}

\paragraph{Rubric Assessment Results.} As shown in Table~\ref{tab:results}, OpenAI Deep Research and Gemini Deep Research significantly outperform other DARS on the rubric assessment, significantly outperforming other DARS with 20-30\% advantages. Perplexity Deep Research achieving moderate performance, while Grok3 DeepSearch \& DeeperSearch demonstrate relatively poor performance. 

\paragraph{Factual Assessment Results.} The factual evaluation reveals a consistent pattern across all systems characterized by high faithfulness but low groundedness, as detailed in Table~\ref{tab:results}. When DARS provide citations, they are generally accurate and well-supported since most of DARS achieved faithfulness score over 0.8, indicating robust source verification mechanisms of DARS. However, the low groundedness scores reveal that substantial portions of generated content lack explicit citation support, suggesting systems rely heavily on internal knowledge or implicit reasoning. Gemini Deep Research achieves the best balance between both metrics, demonstrating its superior citation strategy optimization.

\subsection{Key Findings}

\paragraph{Finding 1: DARS systems outperform LLMs with basic web search capabilities on frontier research tasks.} We observe a substantial performance gap between dedicated DARS and LLMs with Search Tools (e.g., OpenAI Deep Research vs. GPT-4o Search Preview), indicating that models with web search capability alone are insufficient to meet the demands of frontier research questions. However, the performance of Perplexity Sonar Reasoning Pro approaches that of Perplexity Deep Research, even outperforming Grok3 DeepSearch \& DeeperSearch, suggesting that models combining both deep reasoning and web search capabilities can achieve competitive performance to some extent.

\paragraph{Finding 2: Groundedness score and research quality show little correlation in frontier research evaluation.} OpenAI Deep Research, which achieves the best performance in rubric assessment, only attains a low Groundedness score of 0.34. Conversely, Perplexity Sonar Reasoning Pro, which achieves the highest Groundedness score of 0.68, demonstrates mediocre performance in rubric assessment. This inverse relationship suggests that for cutting-edge scientific problems, extensive citation coverage may not necessarily correlate with research quality\cite{aksnes2019citations, uzzi2013atypical}. The low groundedness pattern across high-performing DARS might reflect the unique nature of frontier research evaluation, where valuable insights often emerge from deep synthesis and reasoning processes that transcend direct source attribution.

\begin{figure}[htbp]
\centering
\includegraphics[width=0.8\textwidth]{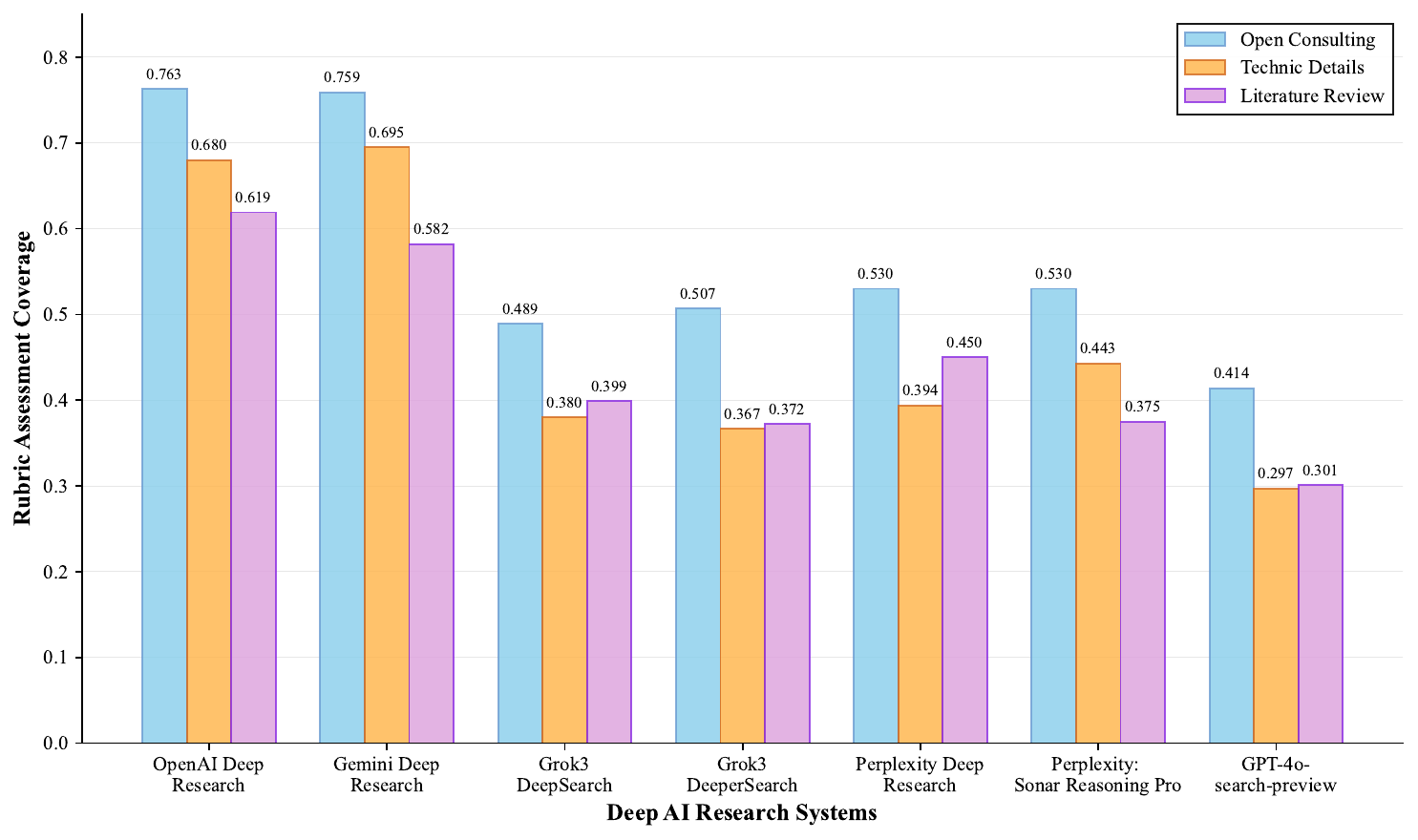}
\caption{Performance Analysis by Question Type (Rubric Assessment Coverage). Performance comparison across different question types for Deep AI Research Systems. Each system shows varying strengths across open consulting, technical details, and literature review categories.}
\label{fig:question_type_performance}
\end{figure}

\paragraph{Finding 3: DARS systems excel at open consulting questions compared to technical details and literature review on ResearcherBench.} Our analysis across different question types reveals distinct capability patterns among DARS systems, as illustrated in Figure~\ref{fig:question_type_performance}. All systems demonstrate better performance on open-ended consulting questions compared to other categories, with top systems achieving 76\%+ coverage rates. Gemini performs best in technical details, while OpenAI leads in open consulting and literature review tasks. The superior performance on Open Consulting questions validates our hypothesis that DARS systems are particularly effective as innovative research ideation partners rather than precision technical implementation guides.

\subsection{Judge Model Selection and Meta Evaluation}
\label{sec: judge_model_sel}

\begin{table}[htbp]
\centering
\begin{tabular}{lccccccccc}
\toprule
\multirow{2}{*}{\textbf{Judge LLM}} & \multicolumn{4}{c}{\textbf{Unweighted}} & \multicolumn{4}{c}{\textbf{Weighted}} & \multirow{2}{*}\centering{\textbf{Avg. Cost (\$)}} \\
\cmidrule(lr){2-5} \cmidrule(lr){6-9}
 & \textbf{Acc.} & \textbf{Prec.} & \textbf{Rec.} & \textbf{F1} & \textbf{Acc.} & \textbf{Prec.} & \textbf{Rec.} & \textbf{F1} & \\
\midrule
DeepSeek R1 & 0.67 & {0.81} & 0.62 & 0.70 & 0.71 & {0.83} & 0.68 & 0.75 & 0.23 \\
Gemini 2.5 Flash & 0.71 & 0.76 & 0.78 & 0.77 & 0.72 & 0.75 & 0.82 & 0.78 & 0.54 \\
GPT-4.1 & 0.71 & 0.75 & \textbf{0.82} & {0.79} & 0.72 & 0.75 & \textbf{0.85} & {0.80} & {0.19} \\
o3 & \textbf{0.76} & 0.80 & {0.81} & \textbf{0.80} & \textbf{0.76} & 0.80 & {0.83} & \textbf{0.81} & 0.22 \\
o3-mini & {0.75} & \textbf{0.85} & 0.74 & {0.79} & \textbf{0.76} & \textbf{0.85} & 0.76 & {0.80} & \textbf{0.13} \\
\bottomrule
\end{tabular}
\caption{Judge LLM Performance Comparison across Evaluation Metrics. Avg. Cost is measured in US dollars (\$). \textbf{Bold values} indicate the best performance for each metric. ``Unweighted'' treats all rubric items equally, while ``Weighted'' considers the assigned importance weights for each rubric item.}
\label{tab:judge_llm_comparison}
\end{table}

To ensure the reliability of our evaluation framework, we systematically compared multiple leading LLMs as judge models to optimize both performance and cost-effectiveness. We conducted a comprehensive meta-evaluation comparing automated rubric assessment with human expert judgments on a validation sample of 10 responses from different questions and different DARS systems. A team of experienced computer science researchers evaluated these responses using the same rubrics employed in our automated assessment, serving as the ground truth for judge model evaluation.

The evaluated models included DeepSeek R1, Gemini-2.5-flash, GPT-4.1, o3, and o3-mini. Our selection criteria encompassed performance consistency (measured by accuracy, precision, recall, and F1-score in agreement with human expert judgments) and cost efficiency (API costs per question) across evaluation tasks. As shown in Table~\ref{tab:judge_llm_comparison}, different models demonstrate varying strengths across evaluation metrics.

Based on these results, we selected \texttt{o3-mini} for rubric assessment due to its optimal balance of consistency (F1-score: 0.80 and precision: 0.85) and cost-effectiveness (\$0.13 per evaluation). For factual assessment, we chose \texttt{GPT-4.1} for its superior performance in long context processing required for comprehensive claim extraction and citation support evaluation.

The strong agreement metrics achieved between human annotators and our selected judge models validate the reliability of our evaluation framework. The weighted F1-score of 0.80 for \texttt{o3-mini} falls within the ``high agreement'' range for AI evaluation versus expert annotation comparisons~\cite{landis1977measurement}, demonstrating that our rubric-based assessment framework effectively captures expert-level evaluation standards while maintaining scalability for comprehensive benchmark evaluation.





\section{Conclusion}

This paper introduces ResearcherBench, the first comprehensive benchmark specifically designed to evaluate Deep AI Research Systems (DARS) on frontier scientific questions. Our work makes several significant contributions to AI research evaluation, including a carefully curated dataset of 65 high-quality frontier research questions sourced from authentic scientific scenarios, spanning 35 AI research subjects and categorized into three distinct types. We introduce a dual evaluation framework consisting of rubric assessment and factual assessment.

By open-sourcing ResearcherBench, we aim to catalyze a new direction in AI research evaluation that prioritizes depth of understanding and insight generation over breadth of information coverage. Our work represents a paradigm shift from evaluating whether DARS can retrieve and summarize information to assessing whether DARS can understand complex problems and provide meaningful insights as genuine research partners. As we progress towards ASI for AI, ResearcherBench provides both a foundation for systematic evaluation and a roadmap for developing AI systems capable of true research partnership in scientific discovery.

\bibliographystyle{acl_natbib}
\bibliography{related}

\appendix
\section{Limitations and Future Works}
\subsection{Limitations}

While ResearcherBench represents a significant advance in evaluating DARS capabilities on frontier research questions, several limitations should be acknowledged.

\paragraph{Domain Specificity.}  Our benchmark focuses exclusively on AI-related research questions, which may limit the generalizability of findings to other scientific domains such as biology, physics, or chemistry. The specialized nature of AI research questions may exhibit different characteristics compared to frontier problems in other fields, and DARS performance patterns might vary across disciplines.

\paragraph{Dataset Scale.} Our dataset size of 65 questions, while carefully curated for quality, represents a relatively small sample that may not capture the full spectrum of frontier research challenges. The distribution across question types (12 technical details, 20 literature review, 33 open consulting) reflects our current understanding of research assistance needs but may not represent the optimal balance for comprehensive evaluation. 

\paragraph{Black-box Commercial Systems.} Our focus on commercial DARS systems limits insights into the fundamental architectural and training approaches that drive performance differences. The black-box nature of these systems prevents deeper analysis of why certain systems excel in specific question types or what design principles enable superior frontier research assistance.

\subsection{Future Works}

Several promising directions emerge from our findings that warrant further investigation. 

\paragraph{Cross-Domain Expansion.} Expanding ResearcherBench to additional scientific domains would provide valuable insights into the domain-generalizability of DARS capabilities. Developing comparable benchmarks for fields such as biology, chemistry, physics, and social sciences would enable cross-domain analysis of research assistance patterns and reveal whether the open consulting superiority we observed generalizes beyond AI research.

\paragraph{Continuous Benchmark Evolution.} As frontier research rapidly evolves, maintaining the relevance and currency of our benchmark requires periodic incorporation of new research questions that reflect the latest developments in AI and related fields. This ongoing evolution will ensure that ResearcherBench remains aligned with the cutting-edge of scientific inquiry, capturing emerging research paradigms and novel challenges that push the boundaries of current knowledge. Regular updates will involve collaboration with active researchers to identify and validate the most pressing and innovative questions in contemporary AI research.

\paragraph{Longitudinal DARS Evaluation.} Implementing systematic longitudinal evaluation of DARS systems would provide crucial insights into capability development trajectories and technological advancement patterns. By continuously assessing new DARS releases and iterations on our benchmark, we can track performance changes over time, identify which aspects of frontier research assistance improve most rapidly, and analyze the developmental pathways of different system architectures. This longitudinal analysis would inform both research priorities and development strategies, while providing the community with empirical evidence of progress in AI research assistance capabilities.

These future directions would collectively advance our understanding of AI research assistance capabilities and contribute to the development of systems that can serve as genuine partners in scientific discovery across diverse domains.

\section{Details in Question Selection}
\label{app:selection_criteria}

\subsection{Question Selection Framework}

To filter high-quality research questions from the initial collection of several hundred research problems for constructing our benchmark, we design a detailed question selection framework. First, since different questions may emphasize different aspects, we use large language models to classify questions into three categories: technical details, literature review, and open consulting (see Table~\ref{tab:q_select}). For these three categories, we set different evaluation criteria respectively. For example, for literature review questions, the stability of research directions and survey scopes is more important, while for open consulting questions, openness and depth may be more crucial.

We quantify these evaluation criteria into rubrics to achieve quantitative scoring of questions on a 1-5 scale. Each question was independently evaluated by at least two experienced computer science researchers using rubrics. Inter-annotator agreement was calculated to ensure consistency, and disagreements were resolved through discussion. Questions achieving average scores of 4.0 or higher across all relevant dimensions were retained for the final benchmark, resulting in our curated set of 65 high-quality frontier research questions.

\subsection{Question Evaluation Rubrics}

\begin{tcolorbox}[
    colback=blue!10,
    colframe=blue!60!black,
    rounded corners,
    title=Technical Details Questions Evaluation Rubric,
    fonttitle=\bfseries\color{white},
    colbacktitle=blue!60!black,
    coltitle=white,
    breakable,
    enhanced,
    break at=-\baselineskip/0pt/\baselineskip  
]

\textbf{Key Evaluation Dimensions:}
\begin{itemize}
    \item \textbf{Technical Specificity}: Whether the question targets specific technical concepts, implementations, or methodologies with clear scope and boundaries.
    \item \textbf{Precision Requirements}: Whether the question demands accurate, detailed explanations that can differentiate between correct and incorrect technical knowledge.
    \item \textbf{Factual Verifiability}: Whether the answer can be verified against authoritative sources (documentation, standards, publications) with objective criteria.
    \item \textbf{Application Value}: Whether the question effectively reflects the model's retrieval capability for fine-grained technical problems.
\end{itemize}

\textbf{Scoring Standards:}
\begin{itemize}
    \item \textbf{5 points}: Question pinpoints specific technical details with clear scope; requires precise, accurate explanations that demonstrate deep technical understanding; answers are easily verifiable against authoritative sources and remain stable over time; excellently tests model's ability to retrieve and explain complex technical knowledge.
    
    \item \textbf{4 points}: Question targets specific technical aspects with good clarity; requires detailed explanations with minor ambiguity; answers are generally verifiable with stable technical foundations; effectively tests technical knowledge retrieval with room for minor improvements.
    
    \item \textbf{3 points}: Question addresses technical details but with moderate specificity; requires explanations that may allow some interpretation; answers are partially verifiable but may depend on context or evolving standards; adequately tests technical knowledge but could be more precise.
    
    \item \textbf{2 points}: Question lacks technical specificity or is too broad; explanations may be vague or surface-level; answers are difficult to verify objectively or depend heavily on time-sensitive information; limited effectiveness in testing precise technical knowledge.
    
    \item \textbf{1 point}: Question is technically vague or overly general; fails to demand specific technical knowledge; answers cannot be objectively verified or are highly dependent on rapidly changing information; ineffective for testing detailed technical understanding.
\end{itemize}

\end{tcolorbox}

\begin{tcolorbox}[
    colback=green!5,
    colframe=green!40!black,
    rounded corners,
    title=Literature Review Questions Evaluation Rubric,
    fonttitle=\bfseries\color{white},
    colbacktitle=green!40!black,
    coltitle=white,
    breakable,
    enhanced,
    break at=-\baselineskip/0pt/\baselineskip  
]

\textbf{Key Evaluation Dimensions:}
\begin{itemize}
    \item \textbf{Research Direction Clarity and Survey Scope}: Whether the question provides clear research direction and can guide comprehensive literature comparison and synthesis across multiple perspectives and methodologies.
    \item \textbf{Literature Coverage Requirements}: Whether the question demands systematic exploration of key papers, major approaches, and important findings in the specified research area.
    \item \textbf{Verifiability and Stability}: Whether the research direction is well-established with literature that is easily retrievable and verifiable, maintaining relevance over time.
\end{itemize}

\textbf{Scoring Standards:}
\begin{itemize}
    \item \textbf{5 points}: Question provides clear research direction with well-defined scope; guides comprehensive literature exploration across multiple dimensions; targets stable research area with abundant, easily accessible literature; excellent potential for meaningful survey output.
    
    \item \textbf{4 points}: Question offers generally clear direction with adequate scope definition; encourages broad literature coverage; targets established research area with good literature accessibility; solid foundation for comprehensive survey work.
    
    \item \textbf{3 points}: Question provides moderate direction clarity with acceptable scope; requires literature coverage but may lack depth requirements; targets reasonably stable research area with moderate literature accessibility; sufficient for basic survey objectives.
    
    \item \textbf{2 points}: Question direction is somewhat unclear or too narrow/broad; limited guidance for comprehensive literature exploration; targets area with limited or hard-to-access literature; marginal value for survey purposes.
    
    \item \textbf{1 point}: Question lacks clear research direction or proper scope definition; fails to guide systematic literature exploration; targets unstable area with poor literature accessibility; inadequate for quality survey work.
\end{itemize}

\end{tcolorbox}

\begin{tcolorbox}[
    colback=orange!10,
    colframe=orange!60!black,
    rounded corners,
    title=Open Consulting Questions Evaluation Rubric,
    fonttitle=\bfseries\color{white},
    colbacktitle=orange!60!black,
    coltitle=white,
    breakable,
    enhanced,
    break at=-\baselineskip/0pt/\baselineskip  
]

\textbf{Key Evaluation Dimensions:}
\begin{itemize}
    \item \textbf{Openness and Depth}: Whether the question encourages creative exploration of multiple perspectives and stimulates deep, multi-dimensional thinking beyond conventional approaches.
    \item \textbf{Forward-Looking Value}: Whether the question addresses emerging trends, future challenges, or strategic insights that provide meaningful guidance for research directions or industry development.
    \item \textbf{Conceptual Innovation Potential}: Whether the question can inspire novel viewpoints, creative problem-solving approaches, or innovative thinking that advance understanding in the field.
    \item \textbf{Balanced Grounding and Long-term Relevance}: Whether the question maintains reasonable connection to existing knowledge while avoiding over-dependence on transient trends, ensuring lasting value.
\end{itemize}

\textbf{Scoring Standards:}
\begin{itemize}
    \item \textbf{5 points}: Question demonstrates exceptional openness that encourages creative exploration across multiple dimensions; addresses significant forward-looking challenges or strategic opportunities; inspires innovative thinking and novel approaches; maintains strong grounding in fundamental principles while offering lasting relevance beyond current trends.
    
    \item \textbf{4 points}: Question shows high openness with good potential for multi-perspective exploration; addresses meaningful future-oriented topics with strategic value; encourages innovative thinking with reasonable grounding; demonstrates good long-term relevance.
    
    \item \textbf{3 points}: Question provides moderate openness with some potential for creative exploration; addresses topics with acceptable forward-looking value; allows for some innovative thinking but may lack depth; maintains reasonable balance between innovation and grounding.
    
    \item \textbf{2 points}: Question offers limited openness with constrained exploration potential; addresses topics with minimal strategic or future value; provides little inspiration for innovative thinking; may be either too abstract or too tied to current trends.
    
    \item \textbf{1 point}: Question lacks meaningful openness and fails to encourage creative exploration; addresses topics with little strategic value or future relevance; provides minimal potential for innovative thinking; either completely ungrounded or overly dependent on temporary trends.
\end{itemize}

\end{tcolorbox}

\section{Guidelines for Rubric Design}
\label{app:rubric_guidelines}

\subsection{Key Insight Extraction} 

We firstly use \texttt{Claude-3.7-Sonnet} to analyze the contextual source material for each question, identifying key insights and generating comprehensive reference materials. For different question types, we design different focus areas when extracting key insights:

\begin{tcolorbox}[
    colback=blue!5,
    colframe=blue!40!black,
    rounded corners,
    title=Technical Details Questions - Key Insight Guidelines,
    fonttitle=\bfseries\color{white},
    colbacktitle=blue!40!black,
    coltitle=white,
    breakable,
    enhanced
]
For technical details questions, the analysis focuses on extracting structured key insights with emphasis on:
\begin{enumerate}
    \item Precise technical specifications and parameters
    \item Detailed algorithmic descriptions and mathematical formulations
    \item Implementation considerations and computational requirements
    \item Performance metrics and efficiency analyses
    \item Technical limitations and edge cases
    \item Optimization techniques and fine-tuning procedures
    \item Code examples or pseudocode where applicable
    \item System architecture and component interactions
    \item Technical dependencies and environmental requirements
    \item Debugging approaches and common technical pitfalls
\end{enumerate}
\end{tcolorbox}

\begin{tcolorbox}[
    colback=green!5,
    colframe=green!40!black,
    rounded corners,
    title=Literature Review Questions - Key Insight Guidelines,
    fonttitle=\bfseries\color{white},
    colbacktitle=green!40!black,
    coltitle=white,
    breakable,
    enhanced
]
For literature review questions, the analysis focuses on extracting structured key insights with emphasis on:
\begin{enumerate}
    \item Comprehensive overview of the technological landscape
    \item Historical development and evolution of relevant technologies
    \item Current state-of-the-art approaches and methodologies
    \item Comparative analysis of different technical solutions
    \item Key research papers, influential publications and bibliographic references
    \item Emerging trends and future research directions
    \item Major contributors and research groups in the field
    \item Theoretical foundations and fundamental principles
    \item Cross-disciplinary connections and applications
    \item Benchmark datasets and evaluation frameworks commonly used in the field
\end{enumerate}
\end{tcolorbox}

\begin{tcolorbox}[
    colback=orange!10,
    colframe=orange!60!black,
    rounded corners,
    title=Open Consulting Questions - Key Insight Guidelines,
    fonttitle=\bfseries\color{white},
    colbacktitle=orange!60!black,
    coltitle=white,
    breakable,
    enhanced
]
For open consulting questions, the analysis focuses on extracting structured key insights with emphasis on:
\begin{enumerate}
    \item Provision of new insights beyond common knowledge or existing literature
    \item In-depth analysis of the question from multiple perspectives
    \item Critical thinking and identification of key challenges and core problems
    \item Novel hypotheses, conceptual frameworks, or alternative viewpoints
    \item Strategic discussions on potential research directions or practical solutions
    \item Integration of cross-disciplinary knowledge to enrich the analysis
    \item Reflection on the broader implications, including societal, ethical, and industrial impacts
    \item Exploration of future trends and transformative opportunities
    \item Expert judgment supported by logical reasoning and evidence
    \item Creative and thought-provoking ideas that inspire further discussion
\end{enumerate}
\end{tcolorbox}

\subsubsection{Key Insight Extraction Prompt}

We employ the following prompt to extract key insights from the contextual source material of questions, and generate auxiliary materials as reference.

\begin{tcolorbox}[
    colback=gray!10,
    colframe=gray!60!black,
    rounded corners,
    title=Key Insight Extraction Prompt Template,
    fonttitle=\bfseries\color{white},
    colbacktitle=gray!60!black,
    coltitle=white,
    breakable,
    enhanced
]
\textbf{\textless system\_role\textgreater }\\
You are an expert research analyst specializing in extracting high-value insights from academic and technical content.\\
\textbf{\textless /system\_role\textgreater }\\

\textbf{\textless user\_prompt\textgreater }\\
Your task is to identify and structure key insights that demonstrate deep understanding and expert-level analysis. Given the following question and its contextual source material, extract key insights following the specific guidelines for \{Question Type\} questions.\\

\textbf{Question:} \{Question\}\\

\textbf{Source Material:} \{Source Context\}\\

\textbf{Guidelines:} \{Guideline for Question Type\}\\

Please identify and extract 8-15 key insights that represent the most valuable and insightful aspects of addressing this question. Each insight should be:
\begin{itemize}
    \item Substantive and demonstrate deep understanding
    \item Directly relevant to answering the question
    \item Represent expert-level analysis or specialized knowledge
    \item Be specific enough to be evaluable
\end{itemize}

Format your response as a structured list of key insights with brief explanations.\\
\textbf{\textless /user\_prompt\textgreater }
\end{tcolorbox}

\begin{tcolorbox}[
    colback=gray!10,
    colframe=gray!60!black,
    rounded corners,
    title=Key Insight Extraction Prompt Template,
    fonttitle=\bfseries\color{white},
    colbacktitle=gray!60!black,
    coltitle=white,
    breakable,
    enhanced
]
\textbf{\textless system\_role\textgreater }\\
You are an expert research analyst specializing in extracting relevant information from the document and providing a comprehensive reference materials. \\
\textbf{\textless /system\_role\textgreater }\\

\textbf{\textless user\_prompt\textgreater }\\
Your task is to:
\begin{itemize}
    \item Carefully analyze the document script provided
    \item Extract all relevant information related to the specific question, containing all the key insights
    \item Organize the information in a coherent, well-structured response
    \item Provide accurate, helpful information based primarily on the document content
\end{itemize}

IMPORTANT: If the document and those key insights do not contain enough information to fully answer the question, you may supplement with your general knowledge, but you must clearly indicate which parts are from the document and which parts are your additional context or expertise. Always prioritize information from the document and provided key insight, and only add relevant knowledge when necessary to provide a more complete answer.\\

\textbf{Question:} \{Question\}\\

\textbf{Full Document:} \{Document\}\\

\textbf{Key Insights:} \{Key Insights List\}\\

Please provide a comprehensive answer to the question based primarily on the information in the document and key insights. If needed, you may supplement the answer with your own knowledge, but clearly distinguish between information from the document and your additional insights.

\textbf{\textless /user\_prompt\textgreater }
\end{tcolorbox}

\vspace{8pt}

\subsubsection{Human Verification and Rubric Design}

Based on the extracted key insights and reference materials, we invite human annotators to design corresponding rubrics for each question. We established the following annotation guidelines:

\begin{tcolorbox}[
    colback=purple!10,
    colframe=purple!60!black,
    rounded corners,
    title=Rubric Design Annotation Guidelines,
    fonttitle=\bfseries\color{white},
    colbacktitle=purple!60!black,
    coltitle=white,
    breakable,
    enhanced
]
\textbf{Task Overview:}\\
You will be presented with a research question, a list of key insights, reference materials, and complete contextual information discussing this research question. Your task is to transform these key insights into a rubric for evaluating the quality of reports answering this research question, and assign weights to different rubric items based on value judgment.\\

\textbf{Rubric Design Requirements:}
\begin{itemize}
    \item \textbf{Clarity and Verifiability}: Each rubric item must be clearly described and easy to verify objectively.
    \item \textbf{Independence}: Ensure each rubric item is independently verifiable without overlap.
    \item \textbf{Conceptual Focus}: Focus on essential concepts rather than specific examples.
    \item \textbf{Evaluative Phrasing}: Phrase each rubric item as an evaluative statement, not a descriptive one.
    \item \textbf{Objective Assessment}: Make rubric items specific enough to enable clear pass/fail evaluation.
    \item \textbf{Action-Oriented Language}: Use verbs like "Explains," "Describes," "Discusses," "Outlines," "Provides," "Analyzes," "Compares," "Identifies" to indicate the expected level of detail and engagement with concepts.
    \item \textbf{Contextual Reference}: The reference materials serve only as auxiliary annotation text and may not accurately reflect the original discussion content. When confused, refer to the original context for specific information.
\end{itemize}

\textbf{Weight Assignment Guidelines:} Assign weights from 1-3 based on the importance of each rubric item to answering the question comprehensively:
\begin{itemize}
    \item Higher weights (3) should be assigned to rubric items that are core to understanding the core question
    \item Medium weights (2) for supporting rubric items that add depth and context
    \item Lower weights (1) for nice-to-have rubric items that enhance but are not essential to the answer
    \item Ensure total weights reflect the relative importance hierarchy of different aspects
\end{itemize}

\textbf{Quality Control Measures:}
\begin{itemize}
    \item Each rubric item should be binary assessable (present/absent)
    \item Avoid subjective language that could lead to inconsistent evaluation
    \item Test each rubric item against the reference materials to ensure it captures meaningful distinctions
    \item Ensure rubric items collectively cover the essential aspects of a comprehensive answer
\end{itemize}
\end{tcolorbox}



\section{Experimental Setup}
\label{app:experiment_setup}

For most DARS systems except Perplexity Deep Research, interaction was only possible through web user interfaces (WebUI), which somewhat limited the scalability of our evaluation experiments. To ensure fairness across different DARS systems, we employed default settings when interacting with all systems. Specifically, OpenAI Deep Research typically asks follow-up questions after the initial user query, and our standard response was "By default." to maintain consistency across evaluations. Gemini-2.5-Pro Deep Research pre-generates research plans before conducting research, and we directly used these generated plans without any modifications to maintain system autonomy and avoid human intervention bias.

For non-DARS systems with web search capabilities (GPT-4o Search Preview and Perplexity: Sonar Reasoning Pro), we configured the Search Context Size to "High" setting to maximize their research capabilities and ensure fair comparison with dedicated research agents. This enhancement was designed to compensate for the lack of specialized research workflows in these systems by providing them with expanded search context, thereby enabling more comprehensive information retrieval and analysis. Additionally, we designed a unified prompt for non-DARS systems to guide models in generating responses with proper report format.

\begin{tcolorbox}[
    colback=gray!10,
    colframe=gray!60!black,
    rounded corners,
    title=Report Generation Prompt Template,
    fonttitle=\bfseries\color{white},
    colbacktitle=gray!60!black,
    coltitle=white,
]
Generate a comprehensive research report addressing the following research questions. Your report must include the clear structure, detailed explanations, and references to relevant academic sources. You should use IEEE style citations for all references, Use numbered citations in square brackets like [1], [2], [3]. 

Research Question: [QUESTION]
\end{tcolorbox}

\section{Data Collection Timeframes}
\label{app:data_coll}

Due to the lack of transparency regarding model iterations and technical details in most commercial DARS systems, we explicitly documented the timeframes during which our data collection occurred. This documentation is crucial for reproducibility and understanding potential temporal effects on system performance.

\begin{table}[htbp]
\centering
\caption{Data Collection Timeframes for DARS}
\label{tab:data_collection_timeframes}
\begin{tabular}{lc}
\toprule
\textbf{DARS} & \textbf{Data Collection Timeframe} \\
\midrule
OpenAI Deep Research & March 24 -- April 29 \\
Perplexity Deep Research & March 24 -- April 15 \\
Grok Deep Search & March 25 -- April 14 \\
Gemini 2.5 Pro Deep Research & April 15 -- April 21 \\
Grok Deeper Search & April 18 -- April 19 \\
\bottomrule
\end{tabular}
\end{table}

\section{Implementation Details and Prompt Templates}
\label{app:implement_details}

\subsection{Implementation Details}

While our evaluation framework follows the methodology outlined in Section \ref{sec:Evaluation_Methodology}, several implementation-specific optimizations were adopted to enhance robustness and efficiency in practice.

\paragraph{Rubric Assessment Implementation.} For rubric assessment, we evaluate each rubric item independently using the prompt template provided in Appendix \ref{prompt:rubric}. 

\paragraph{Factual Assessment Implementation.} For factual assessment, considering computational efficiency and context length limitations, we implemented a section-based claim extraction strategy rather than processing entire reports simultaneously. Specifically, we first segment each DARS-generated report into sections based on paragraph boundaries, then extract claims from each section independently using the prompt template shown in Appendix \ref{prompt:claim_extract}. The extracted claims from all sections are subsequently aggregated to form the complete claim set for each response.

Following claim extraction, we group claims sharing identical URL sources to optimize the verification process and reduce redundant web content retrieval operations. For each URL group, we perform batch Claim Verification Evaluation using the prompt template detailed in Appendix \ref{prompt:claim_judge}. To handle potential extraction failures and web content retrieval errors robustly, we extended the binary judgment framework to include an ``unknown'' category in addition to ``yes'' and ``no'' responses. This three-way classification allows the system to gracefully handle cases where claims are incompletely extracted or source URLs are inaccessible. Claims classified as ``unknown'' are excluded from final metric calculations to ensure evaluation reliability.

\subsection{Rubric Coverage Evaluation Prompt}
\label{prompt:rubric}

\begin{tcolorbox}[
    colback=blue!5,
    colframe=blue!60!black,
    rounded corners,
    title=Rubric Coverage Evaluation Prompt,
    fonttitle=\bfseries\color{white},
    colbacktitle=blue!60!black,
    coltitle=white,
    breakable,
    enhanced
]

\textbf{\#\# Task}\\
Determine whether the AI response adequately covers the specific rubric item provided. Answer with ``yes'' or ``no'' followed by a brief justification.

\textbf{\#\# Input Materials}\\
\textbf{\textless Question\textgreater}: \{question\}\\
\textbf{\textless Rubric Item\textgreater}: \{rubric\}\\
\textbf{\textless Rubric Weight\textgreater}: \{rubric\_weight\} (indicates the importance of this rubric item)\\
\textbf{\textless AI Response\textgreater}: \{ai\_response\}

\textbf{\#\# Evaluation Criteria}
\begin{itemize}
    \item Answer ``yes'' if the AI response clearly includes or adequately expresses the main content of the rubric item
    \item Answer ``yes'' if the response conveys the same meaning as the rubric item, even if using different terminology or phrasing
    \item Answer ``no'' if the AI response only partially addresses or completely fails to mention the content of the rubric item
    \item Consider semantic equivalence, not just keyword matching
    \item Pay special attention to technical details, numerical values, and specific claims in the rubric item
\end{itemize}

\textbf{\#\# Output Format}\\
Your answer must begin with either ``yes'' or ``no'' followed by a brief justification.

\textbf{\#\# Example format}\\
``yes: The response clearly addresses this rubric item by explaining [specific detail]...''\\
``no: While the response mentions [related concept], it fails to address [specific aspect] of the rubric item...''

\end{tcolorbox}

\subsection{Claims Extraction Prompt}
\label{prompt:claim_extract}

\begin{tcolorbox}[
    colback=green!5,
    colframe=green!60!black,
    rounded corners,
    title=Claims Extraction Prompt,
    fonttitle=\bfseries\color{white},
    colbacktitle=green!60!black,
    coltitle=white,
    breakable,
    enhanced
]

\textbf{\#\# Task Description}\\
Extract all factual claims from the provided academic paper. Each claim should be a factual statement that can be verified. Claims may or may not have supporting citations.

\textbf{\#\# Input}\\
A Research Question and a complete academic paper containing factual claims, some of which may have citation markers and corresponding URLs (either inline or in a reference section).

\textbf{\#\# Output Requirements}
\begin{itemize}
    \item Extract each distinct factual claim throughout the entire paper
    \item For each claim, output a JSON object with:
    \begin{itemize}
        \item The exact claim text as a string
        \item The original text from the paper containing this claim (context)
        \item The corresponding citation URL as source (if a citation marker directly follows the claim)
    \end{itemize}
    \item If a claim has a citation marker directly following it, return the supporting URL as source
    \item If a claim does not have a citation marker directly following it, return an empty string for source
    \item Ensure all string values are properly escaped for valid JSON format (e.g. Replace internal quotation marks (") with escaped quotation marks (\textbackslash")) in the claim and context
    \item Return a JSON array containing all claim objects
\end{itemize}

\textbf{\#\# Format Specification}
\begin{verbatim}
[
  {
    "claim": "The exact statement representing a factual claim",
    "context": "The original sentence or passage from the paper 
               containing this claim",
    "source": "https://example.com/source1"
  },
  {
    "claim": "Another factual statement without direct citation",
    "context": "The original sentence or passage from the paper 
               containing this claim",
    "source": ""
  }
]
\end{verbatim}

\textbf{\#\# Guidelines for Claim Identification}
\begin{enumerate}
    \item A claim should be a complete, standalone factual statement
    \item Maintain the original wording where possible, but remove unnecessary context
    \item Extract all factual claims regardless of whether they have citation support
    \item Only consider to map citation markers (numbers, author names, etc.) to their corresponding URLs in the references section when it directly follow the claim statement
    \item Exclude opinions, speculations, or methodological descriptions
    \item Extract the context passage containing each claim for verification purposes
    \item If multiple claims are associated with the same citation, extract them as separate entries
\end{enumerate}

\textbf{\#\# Citation URL Mapping}
\begin{itemize}
    \item If URLs appear directly after claims, use those URLs directly
    \item Citation markers (e.g. follows a number or [number]) must directly follow the claim to be considered as supporting that claim
    \item If claims use citation markers that reference a bibliography or reference section, locate the corresponding URLs in that section
    \item If a claim has no directly following citation marker, use an empty string for source
\end{itemize}

Please extract all claims from the following paper and provide them in the specified JSON format:

Research Question: 
[QUESTION]

Response Content:
[CONTENT]

References:
[REFERENCES]

\end{tcolorbox}

\subsection{Claim Verification Prompt}
\label{prompt:claim_judge}

\begin{tcolorbox}[
    colback=orange!5,
    colframe=orange!60!black,
    rounded corners,
    title=Claim Verification Prompt,
    fonttitle=\bfseries\color{white},
    colbacktitle=orange!60!black,
    coltitle=white,
    breakable,
    enhanced
]

\textbf{\#\# Task Description}\\
Your task is to verify whether multiple claims are supported by the provided reference content.

\textbf{\#\# Input}
\begin{itemize}
    \item A reference content that contains supporting information
    \item A list of claim-context pairs that need to be verified against the reference
\end{itemize}

\textbf{\#\# Output}\\
For each claim, respond with `yes', `no', or `unknown' to indicate whether the claim is supported by the reference content. Output in the specified JSON format.

\textbf{\#\# Output Format Specification}
\begin{verbatim}
[
  {
    "id": 1,
    "result": "yes"
  },
  {
    "id": 2,
    "result": "no"
  },
  {
    "id": 3,
    "result": "unknown"
  }
]
\end{verbatim}

\textbf{\#\# Verification Guidelines}

\textbf{\#\#\# Claim Support Determination}\\
If the reference is valid, for each given claim:
\begin{itemize}
    \item \textbf{`yes'}: If the facts or data in the claim can be found entirely or partially within the reference content
    \item \textbf{`no'}: If all facts and data in the statement cannot be found in the reference content
    \item \textbf{`unknown'}: If verification encounters difficulties (such as semantic incompleteness, ambiguity, or other issues that make verification impossible), or reference contains are not available (`page not found' message, connection errors, or other non-content responses)
\end{itemize}

Notice that claims must be verifiable from the content provided, not based on general knowledge.

\textbf{\#\#\# Using Context Information}\\
If you encounter difficulties when verifying claims (e.g., semantic incompleteness/ambiguity issues), refer to the corresponding additional context. If problems still exist after considering the paragraph context, output `unknown'.

Please provide your verification results in the specified JSON format.

Source:
[SOURCE]

Claim-Context Pair List: 
[CLAIM\_LIST]

\end{tcolorbox}

\end{document}